# Visually Grounded Reasoning across Languages and Cultures

**Fangyu Liu**[*,ð]  **Emanuele Bugliarello**[*,ɕ]  **Edoardo Maria Ponti**[ə,ŋ]
**Siva Reddy**[ə,ŋ]  **Nigel Collier**[ð]  **Desmond Elliott**[ɕ]
[ð]University of Cambridge   [ɕ]University of Copenhagen
[ə]Mila – Quebec Artificial Intelligence Institute   [ŋ]McGill University

## Abstract

The design of widespread vision-and-language datasets and pre-trained encoders directly adopts, or draws inspiration from, the concepts and images of ImageNet. While one can hardly overestimate how much this benchmark contributed to progress in computer vision, it is mostly derived from lexical databases and image queries in English, resulting in source material with a North American or Western European bias. Therefore, we devise a new protocol to construct an ImageNet-style hierarchy representative of more languages and cultures. In particular, we let the selection of both concepts and images be entirely driven by native speakers, rather than scraping them automatically. Specifically, we focus on a typologically diverse set of languages, namely, Indonesian, Mandarin Chinese, Swahili, Tamil, and Turkish. On top of the concepts and images obtained through this new protocol, we create a multilingual dataset for **M**ulticultural **R**easoning over **V**ision and **L**anguage (MaRVL) by eliciting statements from native speaker annotators about pairs of images. The task consists of discriminating whether each grounded statement is true or false. We establish a series of baselines using state-of-the-art models and find that their cross-lingual transfer performance lags dramatically behind supervised performance in English. These results invite us to reassess the robustness and accuracy of current state-of-the-art models beyond a narrow domain, but also open up new exciting challenges for the development of truly multilingual and multicultural systems.

## 1 Introduction

Since its creation, ImageNet (Deng et al., 2009) has charted the course for research in computer vision (Russakovsky et al., 2014). Its backbone consists of a hierarchy of concepts selected from English WordNet (Fellbaum, 2010), a database for lexical semantics. Several other datasets, such as NLVR2 (Suhr

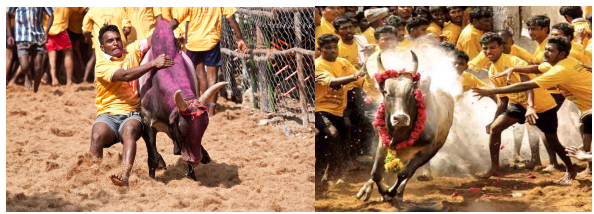

(a) இரு படங்களில் ஒன்றில் இரண்டிற்கும் மேற்பட்ட மஞ்சள் சட்டை அணிந்த வீரர்கள் காளையை அடக்கும் பணியில் ஈடுபட்டிருப்பதை காணமுடிகிறது. ("In one of the two photos, more than two yellow-shirted players are seen engaged in bull taming."). Label: TRUE.

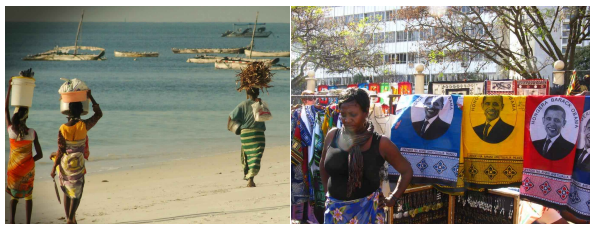

(b) *Picha moja ina watu kadhaa waliovaa nyingine ina leso bila watu.* ("One picture contains several people wearing handkerchiefs and another picture has a handkerchief without people."). Label: FALSE.

Figure 1: Two examples from MaRVL. The Tamil images (a) are from the concept ஏறுதழுவல் (JALLIKATTU, part of an Indian festivity), while the Swahili images (b) are from the concept *leso* (HANDKERCHIEF).

et al., 2019), MS-COCO (Lin et al., 2014), and Visual Genome (Krishna et al., 2017) are built on top of this hierarchy,[1] and likewise many pre-trained encoders of visual data (e.g., ResNet), which are instrumental for transfer learning (Huh et al., 2016).

How suitable are the concepts and images found in ImageNet, beyond the English language and Northern American and European culture in which it was created? Their ideal distribution may be challenging to define and is—to some extent—specific to the intended application (Yang et al., 2020). However, if the goal is world-wide representation, evidence suggests that both the origin (Shankar et al., 2017; de Vries et al., 2019) and content (Stock and Cisse, 2018) of ImageNet data is skewed. To remedy this,

---

*Equal contribution.
Correspondence: fl399@cam.ac.uk, emanuele@di.ku.dk.

[1]Often, only a subset of 1K concepts from the ILSVRC challenges of 2012-2017 (Russakovsky et al., 2014) is considered.

Yang et al. (2020) proposed to intervene on the data, filtering and re-balancing a subset of categories.

Nevertheless, this remains insufficient if the coverage of the original distribution does not encompass multiple languages and cultures in the first place. Hence, to extend the global outreach of multimodal technology, a more radical overhaul of its hierarchy is necessary. In fact, both the most salient concepts (Malt, 1995; Berlin, 2014) and their prototypical members (MacLaury, 1991; Lakoff and Johnson, 1980)—as well as their visual denotations—may vary across cultural or environmental lines. This variation can be obfuscated by common practices in dataset creation, such as (randomly) selecting concepts from language-specific resources, or automatically scraping images from web queries (cf. §2).

In this work, we streamline existing protocols by mitigating the biases they introduce, in order to create multicultural and multilingual datasets. In particular, we let the selection of both concepts and images be driven by members of a community of native speakers. We focus on a diverse set of cultures and languages, including Indonesian, Swahili, Tamil, Turkish, and Mandarin Chinese. In addition, we elicit native-language descriptions by asking annotators to compare and contrast image pairs. The task is to determine whether these grounded descriptions are true or false. We choose this specific task, pioneered by Suhr et al. (2017), as it requires the integration of information across modalities (Goyal et al., 2017) and deep linguistic understanding (Suhr et al., 2019), rather than just matching superficial features (Agrawal et al., 2016). Two examples from our dataset, **M**ulticultu**r**al **R**easoning over **V**ision and **L**anguage (henceforth, MaRVL), are shown in Fig. 1.

We benchmark a series of state-of-the-art visiolinguistic models (Liu et al., 2019; Chen et al., 2020) on MaRVL through both zero-shot and translation-based cross-lingual transfer. Their performance deteriorates considerably compared to an English dataset for the same task (NLVR2; Suhr et al., 2019). Investigating the causes of this failure, we find that while not constructed adversarially, the combined domain shift in concepts, images, and language variety conspires to make MaRVL extremely challenging. Therefore, we conclude, it may provide more reliable estimates of the generalisation abilities of state-of-the-art models compared to existing benchmarks. The dataset, annotation guideline, code and models are available at marvl-challenge.github.io.

## 2 Motivation

The ImageNet Large-Scale Visual Recognition Challenge (Russakovsky et al., 2014, ILSRVC1K) is a landmark evaluation benchmark for computer vision. It is based on a subset of 1,000 concepts from ImageNet (Deng et al., 2009), which consists of a collection of images associated to concepts extracted from the WordNet lexical database (Fellbaum, 2010). This subset is also used as the basis for other multimodal datasets, such as NLVR2 (Suhr et al., 2019). To what extent, however, are images and concepts in these datasets capable of representing multiple languages and cultures? To address this question, we first need to define concepts more precisely.

### 2.1 Concepts: Basic Level and Prototypes

A *concept* is the mental representation of a *category* (e.g. BIRD), where instances of objects and events with similar properties are grouped together (Murphy, 2021). Category members, however, do not all have equal status: some are more *prototypical* than others (e.g. PENGUINS are more *atypical* BIRDS than ROBINS; Rosch, 1973b; Rosch and Mervis, 1975) and boundaries of peripheral members are fuzzy (McCloskey and Glucksberg, 1978; Hampton, 1979). Although prototypes (e.g. for hues and forms) are not entirely arbitrary (Rosch, 1973a), they display a degree of variation across cultures (MacLaury, 1991; Lakoff and Johnson, 1980).

Categories form a hierarchy, from the general to the specific. Among them, *basic-level categories* are cognitively most salient (Rosch et al., 1976). These are used most often to name things by adults (Anglin, 1977) and are the first to be learned by children (Brown, 1958). The basic level, however, is *not* universal, as different cultures may adopt different basic-level concepts (Berlin, 2014). Familiarity of individuals with the domain also plays a role (Wisniewski and Murphy, 1989). So, for instance, in the case of plants, Tzeltal Mayans will choose finer-grained nouns than American undergraduates to indicate the same vine.

Hence, prototypes, basic-level categories, and number of categories for a domain are restricted both by perceptive, cognitive, and environmental factors on the one hand, and by cultural and individual preferences on the other (Malt, 1995; Wierzbicka, 1996).

### 2.2 Limitations of ImageNet

The original annotation of ImageNet was not intended to ensure that its concepts are universal and lie at the basic level (i.e. are most salient for humans); however, these design choices may prove important limitations for the purpose of enabling multi-modal systems to reason over everyday life scenarios in many languages and cultures.

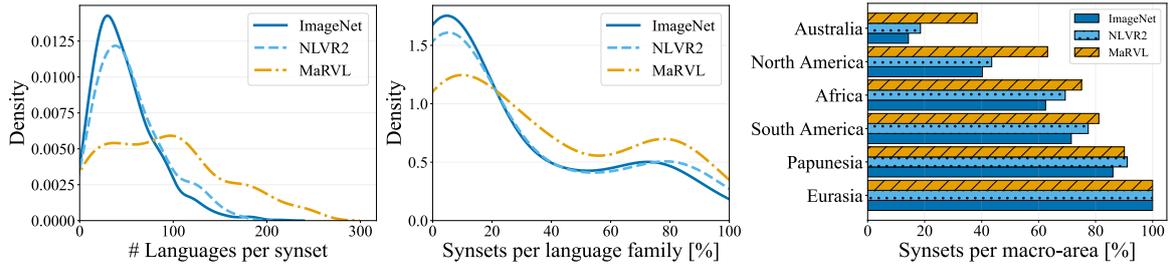

Figure 2: Multilingual and multicultural pervasiveness of ImageNet 1k, NLVR2 and MaRVL synsets by language (left), language family (middle) and macro-area (right) coverage. The Eurasian macro-area is over-represented in ImageNet. Concepts in MaRVL cover more languages and are represented in more language families than the ones in ImageNet.

**ImageNet concepts are not universal.** Since ImageNet is based on the English WordNet, its synsets comprise concepts that are familiar in the Anglosphere but might be exotic or even unknown in other cultures, such as TAILGATING (van Miltenburg et al., 2017). Conversely, it may fail to cover concepts from other cultures (cf. §3.3). To quantify the relevance of the ImageNet concepts across languages, we map each synset manually onto its Wikipedia page[2] and use the Wikipedia API to extract the languages for which a given page is available. Fig. 2 (left) illustrates that most of the synsets are only present in 30 or fewer languages, and that only a small number of "universal" concepts exist. Relying on the WALS database (Dryer and Haspelmath, 2013), we also show that the same argument applies to language families (middle) and that most of these languages are from the Eurasian macro-area (right).

**ImageNet concepts are overly specific to English.** ImageNet contains *overly specific* concepts that belong to leaf nodes in WordNet, e.g. BLENHEIM SPANIEL, rather than *basic-level* concepts such as DOG. To demonstrate this, we calculate the (shortest-path) depth in WordNet of a subset of 447 ImageNet concepts from Ordonez et al. (2013),[3] and compare them with the depth of the labels that people used to refer to objects in corresponding images (Ordonez et al., 2013). Whereas humans tend to employ higher-level synsets (depth $\mu = 8.92$, $\sigma^2 = 3.94$), ImageNet systematically prefers finer-grained synsets ($\mu = 10.61$, $\sigma^2 = 6.13$). Therefore, not only are the concepts overly specific for English, but this mismatch may be aggravated in other cultures. Anecdotally, we found that KOTO, a Japanese instrument, was simply denoted as *instrument* by English speakers; while we expect Japanese annotators to prefer the more precise expression 箏 (*koto*).

### 2.3 Sources of Bias

We now turn our attention to the potential sources of the biases emerging from §2.2. In particular, we scrutinise each individual step that is part of the creation of datasets such as ImageNet, ILSVRC 1K, and NLVR2: 1) Concept selection, 2) Candidate image retrieval, and 3) Manual cleanup.

A first source of bias is the selection of concepts. From WordNet, ImageNet originally selected 12 sub-trees for a total of 5,247 synsets.[4] Finer-grained synsets were preferred to obtain a "*densely populated semantic hierarchy*" (Deng et al., 2009, p. 2). Among these, 1K concepts were selected *randomly* for the ILSVRC 2012-2017 shared tasks.[5] Thus, the 1K concepts form a random sample that may be skewed towards non-basic levels (e.g. 147 synsets are species of dogs).

The second source of bias is candidate image retrieval. Image results from search engines (Flickr and other unspecified engines for ILSVRC 1K; Google Images for NLVR2) do not follow the real-world distribution, e.g. of gender (Kay et al., 2015) and ethnicity (Noble, 2018). In fact, they tend to customise results according to the user's individual profile and localisation. Moreover, queries for ImageNet were again expressed in English, and to a minor (unspecified) extent in Spanish, Dutch, Italian, and Mandarin Chinese, all Western European except for the latter.

Thirdly, additional bias may lie also in image filtering, which is necessary as only an estimate of 10% results per query are of acceptable quality (Torralba et al., 2008). In ImageNet, cleanup was performed via crowdsourcing (in particular, Amazon Mechanical Turk). While no information about the language and culture of the annotators is available, there exists no reason to assume that they were representative of global diversity. Moreover, annotations without consensus were discarded, hence possibly eliminating cultural variations (where disagreement may just represent different basic levels or prototypes).

---

[2] Automatic mapping is hard (Nielsen, 2018).
[3] Please refer to Fig. 5 in App. §B for more details.
[4] Currently expanded to 21,841 synsets according to https://www.image-net.org as of 9 May 2021.
[5] 411 were subsequently substituted to accommodate new challenges (object localisation and fine-grained classification).

## 3 MaRVL: Dataset Annotation

Given the biases in ImageNet-derived or inspired datasets, we define a protocol to collect data that is driven by native speakers of a language, consisting of concepts arising from their lived experiences.[6] The dataset creation consists of five distinct phases: 1) selection of languages; 2) selection of universal concepts; 3) selection of language-specific concepts; 4) selection of images; 5) annotation of captions.

### 3.1 Selection of Languages

We chose five languages, i.e. Indonesian (ID), Swahili (SW), Tamil (TA), Turkish (TR), and Mandarin Chinese (ZH), that are typologically, genealogically, and geographically diverse (Ponti et al., 2020; Clark et al., 2020; Vulić et al., 2020). In addition, this sample covers different writing systems and includes low-resource languages (Tamil and Swahili).[7] This aims at demonstrating the universal applicability of our annotation protocol and reflecting the world's linguistic and cultural diversity during evaluation.

### 3.2 Selection of Universal Concepts

Ideally, datasets for different languages and cultures should reflect the most salient concepts and their typical visual denotations, while retaining some thematic coherence for comparability. Hence, we start from a shared pool of universal semantic fields.

There are multiple lists of human universals from ethnographic studies (Brown, 1991) and from comparative linguistics (Swadesh, 1971; Haspelmath and Tadmor, 2009). As a source for our pool of concepts, we opt for the Intercontinental Dictionary Series (Key and Comrie, 2015), because it is an open-source cooperative and evolving database, and it is cross-lingually comprehensive, as it collects lexical material about languages from all over the world. From its 22 chapters, we retain only the set of 18 semantic fields that cover concrete objects and events. The full list is available in Tab. 6 (App. §A).

### 3.3 Selection of Language-Specific Concepts

For each language, we hire 5 native speaker annotators[8] to provide Wikipedia page links for 5-10 specific concepts in their culture[9] of each semantic field.[10]

The two key requirements are for the concepts to be "commonly seen *or* representative in the speaking population of the language;" and "ideally, to be physical and concrete." For example, for the semantic field MUSIC INSTRUMENT, a Chinese annotator might supply "https://zh.wikipedia.org/wiki/古筝." We reiterated the requirement that the concepts need to be "common/popular" so unusual concepts related to heritage, traditions, folklore, etc. can be avoided. Then, we obtain a ranked list for the most popular concepts in each semantic field and keep only the top 5 concepts (more in case of ties) that have been selected by more than 1 annotator ($> 1$ vote). As a result, we obtain 86–96 specific concepts for each language (see detailed statistics in §4). Among the selected concepts, $72.4\%$ are with $\geq 3$ votes while the remaining $27.6\%$ have 2 votes. The high consensus among annotators suggest that the chosen concepts are representative in the culture.

### 3.4 Selection of Images

After obtaining the full list of concepts, we again hire native annotators to select images for these concepts by collecting image links from the web. We provided a detailed guideline to specify the desired images. In particular, we adopted the image selection requirements of NLVR2: finding images that (1) contain more than one instance of the concept; (2) show an instance of the concept interacting with other objects; (3) show an instance of the concept performing an activity; and (4) display a diverse set of objects or features. These requirements help elicit complex images where the challenges are more likely to lie on compositional reasoning instead of object detection (Suhr et al., 2019). In addition to these, to ensure that the images authentically reflect the native annotators' everyday experience, annotators were also required to select images that are "commonly found or representative in the speaking population of the language." As a result, annotators from different cultures tend to pick visually diverse images even for the same concept, as shown in Fig. 7 (App. §C). Other requirements include images are CC-licensed, natural images, etc. (view App. §D for details). The annotators are allowed to crowd-source the images by any means (local websites, search engines, Wikipedia, etc.) as long as all requirements are met. In general, we made a detailed guideline (available online) to guide the annotators through the whole process and encourage them to find qualified images. We hire two annotators per language. For each concept, we ask for 12 images. Concepts with $< 8$ valid images are discarded.

---

[6] We discard the alternative of selecting concepts from WordNets in other languages (Artale et al., 1997, *inter alia*) as they are often translations from English (thus, bare of culture-specific concepts) and do not specify the basic level for synsets.

[7] All of these languages have large populations of speakers to ensure the availability of sufficient annotators.

[8] We hired 5 speakers per language as we found this was generally enough to achieve a high agreement of selected concepts.

[9] Eliciting the selection from native speakers is ideal as salient concepts are generated first and prototypical members to represent them (such as images) are preferred (cf. §2.1).

[10] In the rare cases where a Wikipedia page is unavailable, the annotators are asked to write the names of the concept directly.

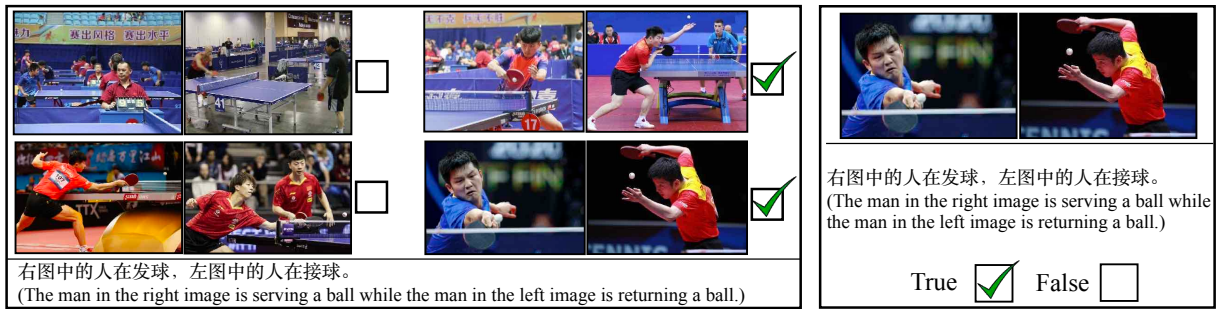

Figure 3: **Left**: For each annotation instance, eight images are randomly picked from the image set of a concept and randomly paired into four pairs. Annotators then write a caption that is True for two pairs but False for the other two. **Right**: Labels are hidden and a different set of annotators will relabel them.

### 3.5 Annotation of Captions

To generate an annotation instance for a given concept, we randomly draw 8 images from its image set, and randomly form 4 image pairs. An annotator is asked to write a caption that is true for two pairs but false for the other two pairs (Fig. 3, left). The captions are required to centre around the "theme concept," which is given during annotation. We do so to prevent annotators from writing captions that rely on oversimplified cues and instead focus on the main objects in the images. Note that each annotation instance contributes 4 data points to our dataset (2 true pairs and 2 false pairs). This annotation scheme largely follows that of NLVR2's (Suhr et al., 2019). We generate 4 annotation instances from each concept.[11] During caption writing, annotators are obliged to report any error (e.g. duplicated images, wrong theme concept name, etc.), and can choose to skip an instance if they find it too hard. We hire native-speaking professional linguists (translators) from proz.com with at least a bachelor's degree to write the captions. Before caption writing, a training session is conducted. For each language, two to four annotators are hired (subject to the availability of annotators in the language).

[11] We aim to collect >1K examples per language, which can be achieved by generating 4 annotation instances per concept.

After a batch of captions is written, we hide the True/False labels and hire a set of validators to relabel all photo-caption pairs (Fig. 3, right). The validators are also required to flag any grammatical errors and typos. After finishing labelling, the "ground-truths" (i.e., labels by the original annotator) are revealed and all conflicted answers are highlighted. Validators then write down why they disagree with the label. The instances with different True/False assignments, along with grammatical errors or typos, are returned to the original annotator for revision. After revisions, the captions are considered finalised. Finally, a native speaker runs a final pass on the examples resolving minor typos and inconsistencies.

In the final dataset, a data point consists of two images, a caption, and a True/False label (Fig. 1).

## 4 Dataset Analysis

**Human validation.** We run a final-round evaluation for reporting human accuracy and inter-annotator agreement (without changing the finalised captions). For each language, 200 examples are randomly sampled from our dataset. We mask the True/False labels and ask two new validators to relabel the examples (same as Fig. 3, right). Across all languages, the Fleiss' kappas among three annotators (one caption writer and two final-round validators) are at least $0.887$ (Tab. 1). According to Landis and Koch (1977), it indicates almost perfect inter-annotator agreement. Suppose the labels given by the caption writer are correct, the average human accu-racy scores[12] are also very high, mostly in the high 90%s except for Swahili (93.0%).

|              | ID   | SW   | ZH   | TR   | TA   | avg. |
|--------------|------|------|------|------|------|------|
| Human accuracy | .963 | .930 | .955 | .970 | .980 | .960 |
| Fleiss' kappa  | .913 | .887 | .933 | .954 | .966 | .931 |

Table 1: Human accuracy of validators on ground-truth labels and inter-annotator agreement (Fleiss' kappa).

|                     | ID   | SW   | ZH   | TR   | TA   | tot. |
|---------------------|------|------|------|------|------|------|
| Concepts selected   | 96   | 88   | 94   | 90   | 86   | 454  |
| with >8 images      | 95   | 78   | 94   | 79   | 83   | 429  |
| % not in WordNet    | 18.8%| 8.0% | 27.7%| 21.1%| 30.2%| 21.1%|
| Total images        | 1153 | 1110 | 1271 | 972  | 946  | 5464 |
| used for captions   | 1091 | 875  | 1107 | 917  | 924  | 4914 |
| Total examples      | 1128 | 1108 | 1012 | 1180 | 1242 | 5670 |
| Total unique captions | 282 | 276 | 253  | 295  | 305  | 1411 |

Table 2: Key stats of concepts and images in MaRVL.

|              | NLVR2 | MaRVL |      |      |      |      |      |
|--------------|-------|-------|------|------|------|------|------|
|              | EN    | ID    | SW   | ZH   | TR   | TA   | avg. |
| Avg. length  | 15.8  | 18.2  | 17.8 | 16.1 | 11.0 | 11.7 | 15.0 |
| Word types   | 811   | 684   | 516  | 848  | 784  | 775  | 721  |
| TTR          | 0.21  | 0.15  | 0.12 | 0.21 | 0.28 | 0.26 | 0.20 |

Table 3: Key stats of MaRVL captions: average length, number of word types, and type-to-token ratio (TTR).

**Concept and image statistics.** For a detailed statistics about our dataset, see Tab. 2. After the image collection, on average 5 concepts are filtered out for each language. Among the final concepts, several are not found in the English WordNet, e.g. sports like *yağlı güreş* (OIL WRESTLING), architectures like 四合院 (SIHEYUAN), or food like தோசை (DOSA).

**Caption statistics.** We report the key statistics of MaRVL captions as well as 250 randomly sampled NLVR2 captions in Tab. 3 The length distributions of the captions are visualised in Fig. 6 (App. §E).

**Image distribution.** To better understand the distribution of MaRVL images (and also how it differs from NLVR2), we extract the features of (1) MaRVL images and (2) 1K randomly sampled NLVR2 images using an ImageNet pre-trained ResNet50 (He et al., 2016), then visualise their embedding distribution using UMAP (McInnes et al., 2018).[13]

As shown in Fig. 4 (top), the Chinese images have very different distributions compared to the English ones (from NLVR2). Specifically, we note that a lot of the clusters of English NLVR2 images are different species of dogs. This is caused by the granularity problem induced by ImageNet. In Fig. 4 (bottom), we compare the image distribution of two languages (Indonesian and Swahili) in MaRVL. Within MaRVL, we still find distributions of images across languages vary. This is largely caused by the distinct concept sets. As shown in the figure, the different clusters stem from the fact that the two regions have very different animal species. We note that, since ResNet50 is pre-trained on ImageNet, the formed clusters may be biased towards ImageNet concepts. As suggested in Fig. 4 (top), the NLVR2 images are usually better clustered than the Chinese ones from MaRVL. UMAP visualisations for more languages are shown in App. §F.

---

[12] We report human accuracy scores as the average correct prediction ratios of the two validators.
[13] We tried running UMAP for multiple times and found the embedding structure generally stable.

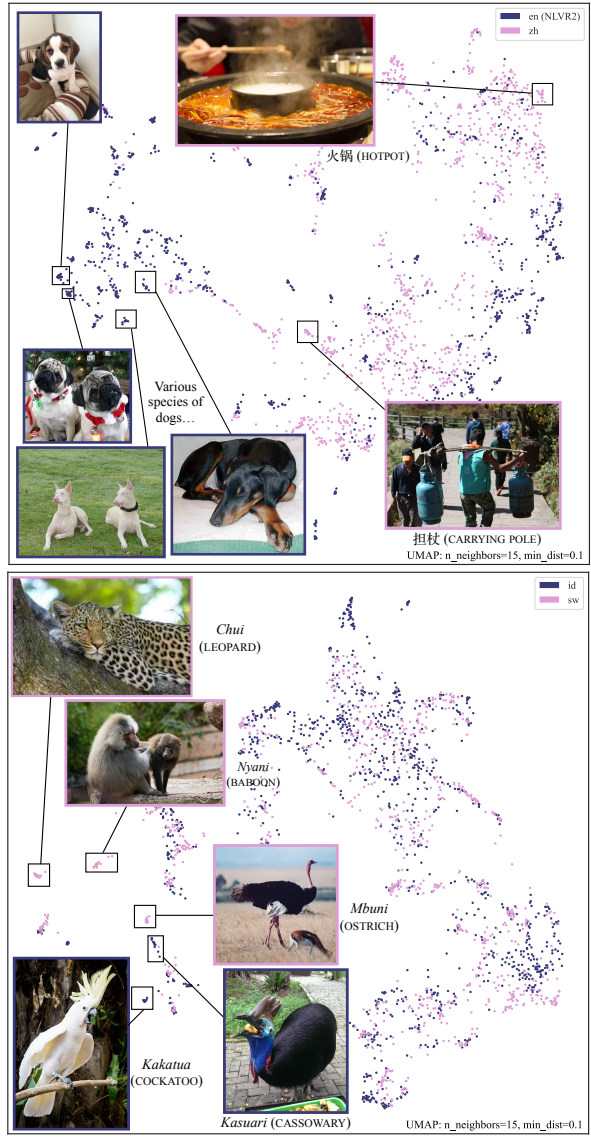

Figure 4: Image feature distributions of MaRVL-ZH and NLVR2 (**top**) and MaRVL-ID and MaRVL-SW (**bottom**).

**Multilingual and multicultural statistics.** Fig. 2 compares key multilingual and multicultural statistics of the concepts in MaRVL with the ones in ImageNet and NLVR2. We can clearly see that MaRVL concepts are present in more languages than the ones in ImageNet and NLVR2 despite being language-specific. We hypothesise this is a result of the concepts in MaRVL being more prototypical and hence reflecting more neighbouring cultures. This is supported by the middle and right plots in Fig. 2 that show how more concepts in MaRVL are found in more language families and macro-areas.

**Limitations.** While we selected the international annotation platforms that cover the most languages (proz.com and prolific.co), it remains challenging to recruit speakers of low-resource languages: we could find only 2-4 qualified annotators per language for caption writing. This might amplify the bias from individual annotators. The authors of the present work do not speak some of the languages considered natively. Moreover, all our concepts are mapped to Wikipedia pages. For low-resource languages, these might be missing for certain concepts. Finally, only ∼5 concepts are selected for each semantic field. This could also inject some bias as frequent concepts are likely to distribute in different categories in an imbalanced way. In general, the protocol for MaRVL can still be improved, albeit our goal to minimise our biases as dataset creators was partly achieved.

## 5 Baselines

Several pre-trained Transformer models for vision-and-language tasks have been proposed, inspired by the BERT architecture (Devlin et al., 2019), and redesigned to handle multimodal inputs. They are pre-trained on large-scale image–text corpora (Sharma et al., 2018), that are usually only available in English. The M³P model (Ni et al., 2021) extends Unicoder-VL (Li et al., 2020a) to encode multilingual inputs, resulting in the first multilingual multimodal BERT-like architecture. Pre-training alternates between modelling multimodal English data and text-only multilingual data. In this paper, we follow this approach and propose two multilingual variants of UNITER (Chen et al., 2020): mUNITER, obtained by initialising UNITER with mBERT (Devlin et al., 2019), and xUNITER, obtained by initialising UNITER with XLM-R$_{BASE}$ (Conneau et al., 2020).

The UNITER architecture consists of a stack of Transformer layers similar to BERT$_{BASE}$, whose input is the concatenation of language and vision embeddings. The language inputs are first split into sub-word units (Wu et al., 2016; Sennrich et al., 2016) and surrounded by two special tokens, $\{[\text{CLS}], w_1, \ldots, w_T, [\text{SEP}]\}$. The language embeddings are then obtained as in the original BERT architecture. The vision input consists of a set of visual features given by a pre-trained object detector, to which we add a special feature [IMG] that encodes the entire image, $\{[\text{IMG}], \mathbf{v}_1, \ldots, \mathbf{v}_K\}$. Each feature is embedded using a BERT-like embedding layer by using its bounding box coordinates as the input position. Finally, the global representation for an image–text pair is obtained via multiplicative pooling (Lu et al., 2019) wherein the pooled representations for the text modality, extracted from the [CLS] token, and for the visual modality, extracted from the [IMG] feature, are element-wise multiplied to obtain a single vector for the image–text pair.

We code our models in VOLTA[14] and pre-train them using the same data and hyperparameters as in the *controlled* setup proposed by Bugliarello et al. (2021).[15] This lets us fairly compare the performance of our multilingual versions with the corresponding monolingual ones. Afterwards, our models are fine-tuned on NLVR2 following the approach initially proposed by Lu et al. (2020). For more details regarding our architecture, pre-training and fine-tuning, see App. §G. After English fine-tuning, multilingual models are tested on MaRVL in a '*zero-shot*' cross-lingual transfer setting.

In addition, we also benchmark the performance of five *monolingual* vision-and-language BERT models available in VOLTA: UNITER, VL-BERT (Su et al., 2020), VisualBERT (Li et al., 2019a), ViL-BERT (Lu et al., 2019) and LXMERT (Tan and Bansal, 2019). These models are also pre-trained in the same controlled setup and fine-tuned on the English training set of NLVR2. Following the established '*translate test*' approach to cross-lingual transfer (Banea et al., 2008; Conneau et al., 2018; Ponti et al., 2021b), they are evaluated on the test sets of MaRVL automatically translated into English.[16]

## 6 Results

In Tab. 4, we show the performance of the baselines outlined in §5 on MaRVL. We report two metrics: **accuracy** across all examples and **consistency**, the proportion of unique sentences for which predictions on all corresponding image pairs are correct. We note that the differences among all models for specific transfer methods are *not* statistically significant. This indicates that varying neural architectures does not have an appreciable impact on performance once they are pre-trained on the same amount of data.

---
[14]https://github.com/e-bug/volta.
[15]As multilingual data, we use Wikipedia in 104 languages.
[16]We use neural machine translation in the Google Cloud API.

|  | NLVR2 | MaRVL | | | | | |
|---|---|---|---|---|---|---|---|
|  | EN | ZH | TA | SW | ID | TR | avg. |
|  | | *Zero-shot* | | | | | |
| mUNITER | 73.2 / 37.7 | 56.8 / 6.7 | 52.2 / 2.3 | 51.5 / 4.7 | 55.0 / 7.8 | 54.7 / 6.8 | 54.0 / 5.7 |
| xUNITER | 72.8 / 35.6 | 55.0 / 5.9 | 55.1 / 7.2 | 55.5 / 5.1 | 57.1 / 9.6 | 58.0 / 8.5 | 56.1 / 7.3 |
|  | | *Translate test* | | | | | |
| LXMERT | 71.1 / 34.1 | 59.4 / 11.2 | 61.9 / 15.4 | 62.6 / 16.5 | 61.4 / 12.4 | 65.7 / 22.0 | 62.2 / 15.5 |
| ViLBERT | 72.2 / 35.9 | 58.4 / 13.2 | **64.3** / 17.5 | 63.8 / 15.7 | 60.7 / 11.7 | **70.4** / 25.1 | 63.5 / 16.6 |
| VisualBERT | 72.7 / 35.4 | 58.5 / 13.5 | 60.6 / 11.6 | 62.5 / 15.4 | 59.7 / 10.6 | 69.4 / 24.7 | 62.1 / 15.2 |
| VL-BERT | 73.3 / 37.1 | 62.8 / 14.7 | 62.5 / **19.2** | **65.8** / **20.6** | 61.0 / **14.9** | 70.3 / **27.5** | **64.5** / **19.4** |
| UNITER | **73.7** / **38.2** | 62.8 / 17.1 | 63.5 / 18.2 | 61.9 / 13.9 | 61.6 / 14.5 | 70.3 / 26.1 | 64.0 / 18.0 |
| mUNITER | 73.2 / 37.7 | 62.7 / **17.9** | 62.3 / **19.2** | 63.4 / 16.5 | 59.8 / 11.4 | 69.2 / 27.1 | 63.5 / 18.4 |
| xUNITER | 72.8 / 35.6 | **63.3** / 16.7 | 62.4 / 16.1 | 64.1 / 15.0 | **62.4** / **14.9** | 69.8 / 16.7 | 64.4 / 17.9 |

Table 4: Performance (accuracy/consistency) in MaRVL and NLVR2 (Test-P). Translate test denotes the setup of the multilingual MaRVL datasets translated into English. Best scores are put in bold, but do *not* imply statistical significance.

| Model | MaRVL ZH→EN | NLVR2₁ₖ EN | NLVR2₁ₖ EN→ZH |
|---|---|---|---|
| LXMERT | 61.7 / 14.3 | - | - |
| ViLBERT | 62.5 / 13.5 | - | - |
| VisualBERT | 59.8 / 13.1 | - | - |
| VL-BERT | 65.4 / 20.6 | - | - |
| UNITER | 63.8 / 19.4 | - | - |
| mUNITER | 61.3 / 17.1 | 72.2 / 33.2 | 56.2 / 6.8 |
| xUNITER | 64.4 / 20.6 | 73.0 / 31.6 | 57.1 / 9.2 |

Table 5: Performance (accuracy/consistency) in MaRVL-ZH when manually translated into English and NLVR2 1k when manually translated into Mandarin Chinese.

**Zero-shot vs. translate test.** We find that both multilingual and monolingual models perform comparably well in English (NLVR2). When evaluated on the languages in MaRVL, however, the performance of zero-shot multilingual baselines dramatically drops by $10-20\%$ points, floating just above chance level. Remarkably, this is also the case for resource-rich languages like Mandarin Chinese (ZH), for which unlabelled text is abundant. Compared to zero-shot transfer, all translate-test baselines gain $4-15\%$ across the different languages, with Turkish improving the most. Yet, there persists a considerable gap of more than $10\%$ compared to the performance on English in NLVR2. Arguably, this is caused by the out-of-distribution nature of the data in MaRVL.

**Disentangling shifts in distribution.** There are two sources of difficulty that make MaRVL challenging: 1) *cross-lingual transfer* and 2) *out-of-distribution* concepts, in both images and descriptions, with respect to English datasets. To assess the effect of each of these factors on model performance, we conduct a controlled study on Mandarin Chinese (MaRVL-ZH).

First, we manually translate MaRVL-ZH into English, hence removing any possible confound due to machine translation in Tab. 4. As shown in Tab. 5 (left column), compared to the *Translate test* evaluation, each model improves its accuracy by only 1-2%, with the exception of mUNITER. Thus, the translation is fairly reliable. Moreover, out-of-distribution concepts are responsible for the largest share of errors (on average, a drop in accuracy of $10\%$).

Second, we sample 250 unique descriptions, corresponding to 1,000 data points, from the NLVR2 test set and manually translate them into Mandarin Chinese. The performance of mUNITER and xUNITER on this subset, which we denote as NLVR2₁ₖ, is listed in Tab. 5 (right column). Although all data points are in-domain, both our multilingual models, mUNITER and xUNITER, lose $16\%$ in accuracy compared to the English NLVR2 1k test set (central column). Hence, this gap can be attributed to cross-lingual transfer from English into Chinese.

**Translate train.** Finally, we establish a baseline for a third possible approach to cross-lingual transfer, 'translate train'. To this end, we machine translate the training set of NLVR2 into Mandarin Chinese and then evaluate on MaRVL-ZH. We find that the performance of mUNITER (62.5/18.7) and xUNITER (61.8/16.7) is close to their respective performance when machine translating MaRVL-ZH into English (*translate test*). Again, the lack of access to culturally relevant concepts hinders generalisation.

## 7 Related Work

**Grounded language reasoning.** Several datasets to assess reasoning over language and vision have been released in the recent years (Antol et al., 2015; Kazemzadeh et al., 2014; Xie et al., 2019; Zellers et al., 2019). Most notably, CLEVR and its extensions (Johnson et al., 2017a,b; Liu et al., 2019; Kottur et al., 2019), as well as GQA (Hudson and Manning, 2019), address the task of answering complex, compositional questions over images. NLVR (Suhr et al., 2017), on the other hand, compares and contrasts pairs of synthetic images via human-written descriptions. The reasoning capabilities of artificial models are evaluated through binary classification over these grounded descriptions. We closely follow the task formulation of NLVR2 (Suhr et al., 2019), which extends the NLVR collection to real-world photographs. Our dataset, MaRVL, is the first multilingual and multicultural dataset for grounded language reasoning.

**Multilingual multimodal datasets.** Elliott et al. (2016) introduced Multi30k, one of the most widely used multilingual image-caption datasets. Multi30k enriches Flickr30K (Young et al., 2014)—originally in English—with translated and newly written German captions. Others provided captions for subsets of MS-COCO images in German and French (Rajendran et al., 2016), Japanese (Yoshikawa et al., 2017), and Chinese (Li et al., 2019b). Chinese captions were made available also for videos by Wang et al. (2019). All these datasets suffer from at least one of these two deficiencies: 1) they start from a sample of images crowd-sourced from North America and Western Europe; 2) they contain at most three (high-resource) languages, only a pale reflection of the world's cross-lingual and cross-cultural variation. Finally, Srinivasan et al. (2021) automatically scraped a multilingual text–image dataset from Wikipedia. While not suitable for evaluation, this is a promising training resource for multilingual multimodal models.

## 8 Conclusions and Future Work

Our empirical and theoretical analyses reveal that concepts and images documented in existing visiolinguistic datasets are likely neither salient nor prototypical in many languages different from English and in cultures outside Europe and North America. Therefore, to mitigate these biases, we devise a new annotation protocol where the selection of images and captions is entirely driven by native speakers. Moreover, we elicit descriptions comparing and contrasting image pairs in 5 typologically diverse languages: Indonesian, Mandarin Chinese, Swahili, Tamil, and Turkish. We publicly release the resulting multicultural and multilingual dataset for grounded language reasoning, MaRVL, and its annotation guidelines, with the hope that other members of the scientific community will be able to further expand it.

Moreover, we develop and benchmark a series of multilingual and multimodal baselines, including model- and translation-based transfer. We find that their performance is sometimes just above chance level and suffers considerably from the out-of-distribution nature of concepts, images, and languages in MaRVL compared to English datasets. This gives us reason to believe that it offers a more faithful estimate of state-of-the-art models' suitability in real-world applications, outside a narrow linguistic and cultural domain.

Better methods for out-of-distribution cross-lingual transfer (Ponti et al., 2021a) and a deeper understanding of how visual stimuli affect language (Li et al., 2020b; Rodríguez Luna et al., 2020) might prove essential for future progress in the MaRVL challenge. In future work, we will also assess model performance on additional tasks already supported by our dataset, such as object recognition, and experiment with multilingual extensions of visiolinguistic models based on contrastive learning (Carlsson and Ekgren, 2021).

## Acknowledgements

🇪🇺 We are grateful to the anonymous reviewers, Rita Ramos, Ákos Kádár, Stella Frank, and members of the CoAStaL NLP group for their constructive feedback. We also thank Alane Suhr for clarifications regarding the NLVR2 dataset collection protocol. This project has received funding from the European Union's Horizon 2020 research and innovation programme under the Marie Skłodowska-Curie grant agreement No 801199, and Cambridge Digital Humanities Digitisation/Digital Resources Awards.

## Ethics Statement

We collected data from crowdworkers spread across the world in languages that are under-represented in current vision and language datasets. Our reasoning was two-fold: first, we believe that researchers should work with lower-resourced languages, and second, we believe that language data should be drawn from different language families.

We hired workers from two crowdsourcing platforms: Prolific and Proz. The identity of the workers on Prolific were anonymous to us, while those on Proz were not anonymous. The use of Profilic was authorised by the relevant authority in our university.

Workers hired from Prolific were predominantly based in North American and Western European countries; those hired on Proz were nearly exclusively based in Indonesia, India, Turkey, China, Tanzania, Kenya, and Somalia. All workers were paid £15–£20/hour, and were paid even if their data did not appear in the final datasets. (There were several rounds of filtering in the collection process.)

## A Chapter to Semantic Field Mapping

We list all chapters, semantic fields, and the mapping between them in Tab. 6.

## B Depth of ImageNet Concepts in the WordNet Hierarchy

Fig. 5 compares the distribution of minimum and maximum depth in WordNet for 447 synsets from Ordonez et al. (2013) according to their labels in ImageNet and to the labels given by the annotators (i.e. basic-level categories). The mapping of human labels onto WordNet was done manually, resolving any disambiguation. We find that humans typically tend to prefer higher-level synsets than the ones present in ImageNet, showing that concepts in ImageNet are already over-specific for English.

## C BASKETBALL across Languages

Fig. 7 shows images for the concept BASKETBALL across our five languages in MaRVL.

## D Additional Annotation Details

**Selection of language-specific concepts.** We pay 0.1 GBP for each Wikipedia link (maximum 10 links for each semantic field) such that annotators would have incentives to write down as many concepts as possible for each semantic field. Annotators hired in this phase are from prolific.co[17] and proz.com.

**Selection of images.** We have also included requirements such as avoid (1) synthetic images (2) collages (3) watermarks (4) low-resolution images, etc. to make sure that only natural images (photos) are included in our dataset. CC-license is also demanded for all collected images. For each valid image, we pay 0.12 GBP. After collecting enough images, we run a second round check by ourselves to delete all unqualified images. Annotators hired in this phase are from prolific.co and proz.com.

**Annotation of descriptions.** Before assigning the task, we conduct training sessions with the description writers. They are required to complete a sample. We then provide feedback and ask for revisions if the sample is not perfectly aligned with the guideline. Each valid description is paid with 0.6 GBP. During quality control, 0.1 GBP is paid for every reviewed example. Description writers are all hired from proz.com while validators are from both proz.com and prolific.co.

---
[17] An extra 33.3% VAT is paid to the platform for all prolific.co payment.

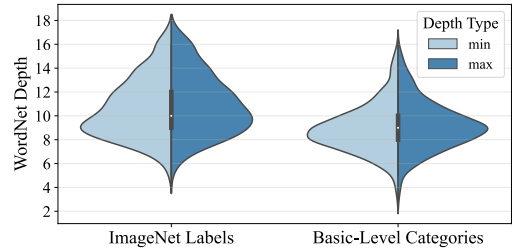

Figure 5: Distributions of minimum and maximum synset depth in the WordNet hierarchy for 447 ImageNet labels compared to basic-level categories provided by humans (Ordonez et al., 2013). The lower depth of the latter suggests that ImageNet categories may be over-specific.

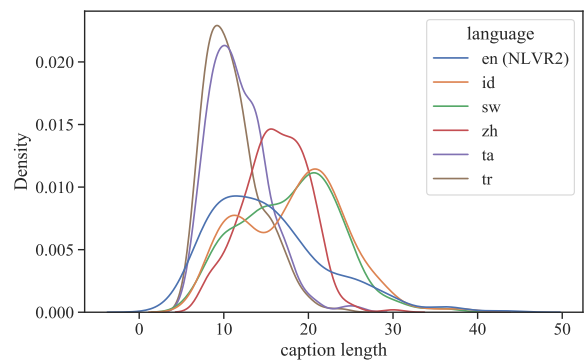

Figure 6: Description length distribution by languages.

**Final-round human validation.** In this round, the re-labelling annotators are asked to focus on logical correctness instead of the grammaticality/fluency of the language. Again, 0.1 GBP is paid for every assigned label. Validators are from both proz.com and prolific.co.

## E Additional Dataset Statistics

We plot the distribution of the description lengths in Fig. 6. Except for Turkish and Tamil, our collected descriptions are longer than the ones in NLVR2.

## F More Image Embedding Visualisations

We plotted image feature distributions of MaRVL-ZH, SW, ID in the main text. Here we also visualise the features of TA, TR (Fig. 8, left), and provide a full graph for all languages (Fig. 8, right).

## G Baselines Details

In this section, we describe our multilingual vision-and-language models, mUNITER and xUNITER, as well as their training procedures in detail. We use the same hyperparameters as in the controlled study

| Chapter | Index | Semantic field |
|---|---|---|
| Animal | 3 | Bird, mammal |
| Food and Beverages | 5 | Food, Beverages |
| Clothing and grooming | 6 | Clothing |
| The house | 7 | Interior, exterior |
| Agriculture and vegetation | 8 | Flower, fruit, vegetable, agriculture |
| Basic actions and technology | 9 | Utensil/tool |
| Motion | 10 | Sport |
| Time | 14 | Celebrations |
| Cognition | 17 | Education |
| Speech and language | 18 | Music (instruments), visual arts |
| Religion and belief | 19 | Religion |

Table 6: Chapter to semantic field mapping.

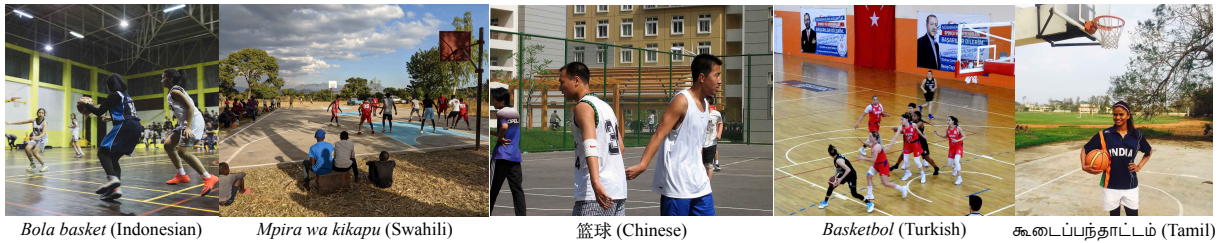

*Bola basket* (Indonesian)    *Mpira wa kikapu* (Swahili)    篮球 (Chinese)    *Basketbol* (Turkish)    கூடைப்பந்தாட்டம் (Tamil)

Figure 7: Five examples of "basketball" from our dataset (one per language). Though describing the same concept, the visual representations can vary drastically across languages/cultures, e.g. in the personal attributes of the players or the field background.

of Bugliarello et al. (2021) and build our models on top of the VOLTA repository.[18]

**Architecture.** Our models extend the UNITER (Chen et al., 2020) architecture. This is a single-stream system consisting of a single stack of Transformer layers, similar to BERT (Devlin et al., 2019). The input to the model is the concatenation of language and vision inputs. The text input consists of a sequence of sub-word units (Wu et al., 2016; Sennrich et al., 2016) surrounded by two special tokens, $\{[\text{CLS}], w_1, \ldots, w_T, [\text{SEP}]\}$. The vision input consists of a set of visual features given by a pre-trained object detector and a special feature [IMG] that encodes the entire image, $\{[\text{IMG}], \mathbf{v}_1, \ldots, \mathbf{v}_K\}$. Specifically, we use $K = 36$ visual features per image given by a Faster R-CNN architecture (Anderson et al., 2018). Similar to BERT$_{\text{BASE}}$, the body of our Transformers consists of 12 layers, each with 12 attention heads, and a hidden size with dimension 768. Finally, the global representation for an image–text pair is obtained via multiplicative pooling (Lu et al., 2019) wherein the pooled representations for the text modality, extracted from the [CLS] token, and for the visual modality, extracted from the [IMG] feature, are element-wise multiplied to obtain a single vector for the image–text pair.

**Pre-training.** Following Ni et al. (2021), we pre-train our models by alternating multilingual text-only batches and multimodal English-only batches. Masked language modelling (MLM; Devlin et al. 2019) is the sole objective used in the multilingual steps, while the loss in multimodal batches is given by the sum of three objectives: MLM, masked region classification with KL-divergence (MRC-KL; Lu et al. 2019) and image–text matching (ITM; Chen et al. 2020). During multimodal pre-training, some of the image regions are randomly masked[19] and the model is tasked to predict the distribution of the corresponding classes given by the Faster R-CNN model. ITM is the multimodal version of next sentence prediction in BERT. Here, the caption of an image is replaced with probability $0.5$ with a random caption in the training corpus. The model is then trained to identify whether the given image–caption pairs are a match or not. Following Lu et al. (2020), MLM and MRC-KL losses are not computed when image and captions are not matched to avoid sub-optimal mapping between visual and linguistic inputs. For multilingual pre-training, we use Wikipedia dumps (Rosa,

---
[18]https://github.com/e-bug/volta.

[19]Regions whose IoU is greater than 0.4 are also masked.

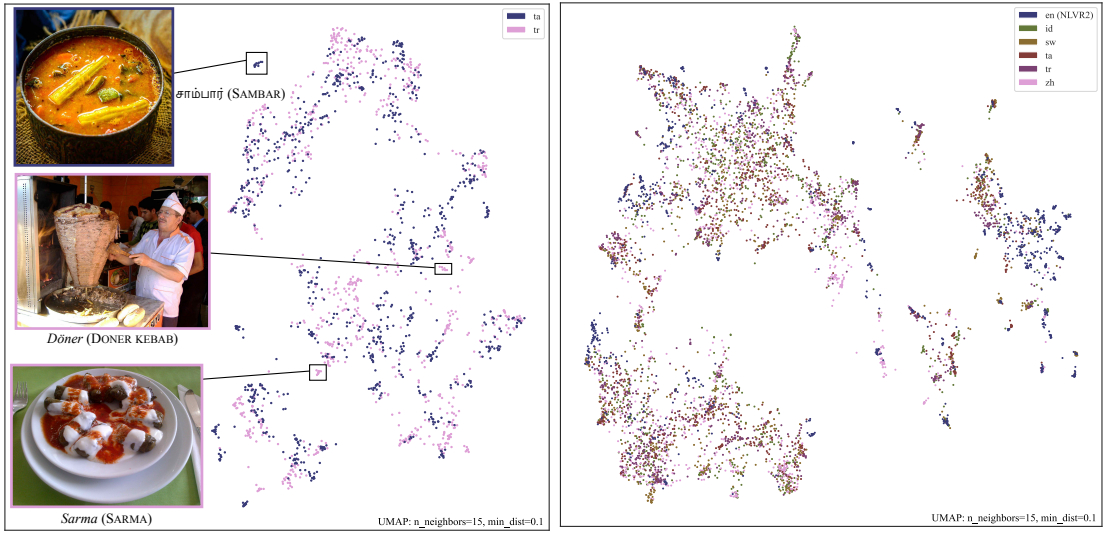

Figure 8: **Left:** image feature distributions of MaRVL-TA, TR. **Right:** image feature distributions of all languages in MaRVL and NLVR2.

|  | mUNITER | | | | | xUNITER | | | | |
|---|---|---|---|---|---|---|---|---|---|---|
| Animal | 53.7 | 57.8 | 55.3 | 57.5 | 58.1 | 53.7 | 58.9 | 61.8 | 65.6 | 52.2 |
| Food and Beverages | 50.0 | 46.5 | 59.0 | 55.3 | 58.1 | 58.3 | 57.6 | 53.8 | 63.2 | 54.4 |
| Clothing and grooming | 59.7 | 50.0 | 56.2 | 56.6 | 70.3 | 66.7 | 55.3 | 62.5 | 53.9 | 54.7 |
| The house | 63.3 | 51.0 | 51.4 | 50.0 | 65.0 | 56.7 | 59.4 | 63.9 | 55.1 | 57.5 |
| Agriculture and vegetation | 52.5 | 48.5 | 49.0 | 50.9 | 54.1 | 52.9 | 51.5 | 53.5 | 59.6 | 56.1 |
| Basic actions and technology | 55.9 | 53.1 | 43.4 | 56.2 | 41.7 | 60.3 | 53.1 | 51.3 | 47.5 | 50.0 |
| Motion | 53.6 | 62.5 | 50.0 | 61.8 | 61.7 | 59.5 | 37.5 | 42.3 | 50.0 | 55.0 |
| Time | 51.7 | 52.8 | 50.0 | 46.7 | 58.3 | 56.7 | 50.0 | 47.5 | 61.7 | 55.0 |
| Cognition | 55.0 | 48.6 | 50.0 | 63.9 | 44.6 | 51.7 | 54.2 | 53.1 | 55.6 | 58.9 |
| Speech and language | 58.3 | 50.9 | 53.3 | 53.2 | 53.8 | 54.6 | 59.3 | 52.5 | 54.8 | 55.8 |
| Religion and belief | 55.0 | 60.7 | 62.5 | 68.8 | 62.5 | 71.7 | 57.1 | 57.5 | 62.5 | 54.2 |
| ALL | 55.1 | 51.5 | 52.3 | 54.7 | 56.8 | 57.1 | 55.5 | 55.2 | 58.0 | 55.0 |
|  | id | sw | ta | tr | zh | id | sw | ta | tr | zh |

Figure 9: Accuracy of mUNITER and xUNITER when grouping concepts by chapter. 'ALL' denotes the overall accuracy.

2018) for 104 languages,[20] while we rely on Conceptual Captions for multimodal pre-training.[21] In the multilingual steps, we first sample languages according to the following multinomial distribution that accounts for the size of each Wikipedia dump (Conneau and Lample, 2019), and then, for each sampled language, a sentence is uniformly sampled. We set the multinomial parameter $\alpha$ as in the original papers: 0.7 for mBERT-based mUNITER and 0.3 for XLM-R-based xUNITER. Each model is pre-trained for 10 Conceptual Captions epochs as done in previous work. Following the implementation of Ni et al. (2021), the models' parameters are updated after each multilingual and multimodal batch, while the learning rate scheduler only after both of them.

[20]https://github.com/google-research/bert/blob/master/multilingual.md
[21]We use the 2.77M data points available in VOLTA.

**Fine-tuning.** We fine-tune our models on the English NLVR2 dataset and then measure zero-shot performance on our MaRVL dataset. In NLVR2, given two images ($I_l$ and $I_r$) and a description $D$, the model is trained to assess the validity of the description given the images (true or false for both images). We follow Lu et al. (2020) and cast this as a classification problem: Given embeddings that encode the two image–description pairs, $(I_l, D)$ and $(I_r, D)$, the probability that they are both valid is predicted by a 2-layer MLP with a GeLU (Hendrycks and Gimpel, 2016) activation in between, followed by a softmax over two classes (representing true and false labels):

$$\mathbb{P}(C|I_l, I_r, D) = \text{softmax}\left(\text{MLP}\left(\begin{bmatrix} h^l_{[\text{IMG}]} \odot h^l_{[\text{CLS}]} \\ h^r_{[\text{IMG}]} \odot h^r_{[\text{CLS}]} \end{bmatrix}\right)\right), \quad (1)$$

where [ ] denotes the concatenation of the pooled representations.

**Experimental setup.** We train our models on a single NVIDIA Titan RTX. Pre-training each model takes 9 days, while NLVR2 fine-tuning for 20 epochs takes 1 day. The parameter sets giving the best validation performance in NLVR2 are used for zero-shot, cross-lingual evaluation on MaRVL.

## H Performance by Chapter

Fig. 9 shows the performance of mUNITER and xUNITER for each chapter in each language. Compared to the overall performance ("ALL"), we find that no chapter is easier or difficult in all languages for both models. Cross-lingual performance of the two models varies acriss chapters: While both models find smaller fluctuations on "Speech and language", mUNITER shows minimal deviation on "Agriculture and vegetation" but twice as much as xUNITER's on "Cognition." Interestingly, in Swahili, mUNITER and xUNITER show opposite behaviour for "Motion," with mUNITER largely outperforming the overall accuracy, while a similarly large drop is given by xUNITER. Overall, we find that both models find almost every chapter almost as difficult, resulting in per-chapter accuracy close to the overall one.

## I More Examples from MaRVL

To demonstrate more details of our dataset, we pick two examples from each language (TA: Fig. 10, ZH: Fig. 11, SW: Fig. 12, ID: Fig. 13 TR: Fig. 14).

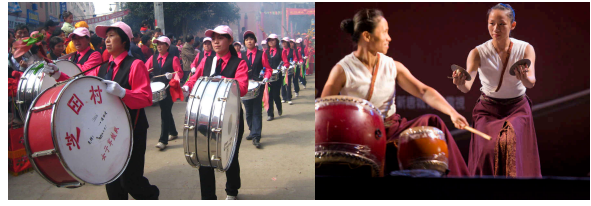

(a) 两张图加起来总共超过五个人在打鼓，并且两张图的人所打鼓的种类不同。("In total, there are more than people playing drums in the two images combined and people in the two images are playing different kinds of drums.", concept: 鼓 (DRUM), label: TRUE)

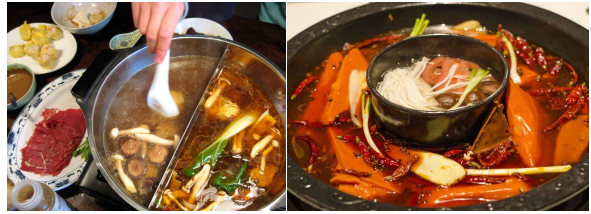

(b) 两图中至少有一张图里面是一口鴛鴦鍋。("At least one of the two pictures shows a mandarin duck pot.", concept: 鴛鴦鍋 (MANDARIN DUCK POT, a specific type of pot), label: TRUE)

Figure 11: More examples from MaRVL-ZH.

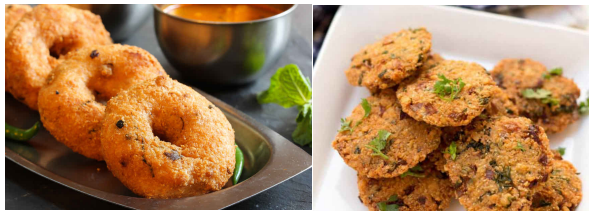

(a) இரண்டு ... மசால் வடைகள் உள்ளான. ("Both ... sala vadas.", concept: வடை (VADA, a popular Indian food), label: FALSE)

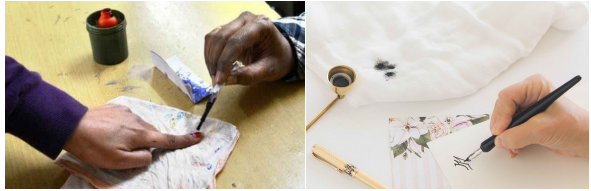

(b) இரண்டு படங்களில் ஒன்றில் காண்பிக்கப்படும் விரலில் வாக்கு மையை காண முடிகிறது. ("In one of the two pictures, the finger shows the vote.", concept: மை (INK), label: TRUE)

Figure 10: More examples from MaRVL-TA.

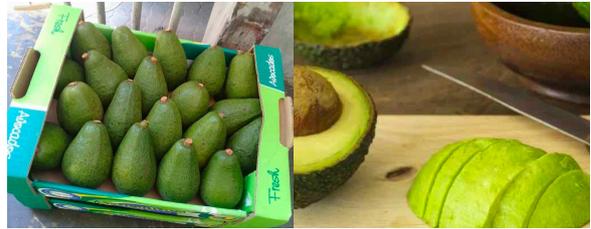

(a) *Picha ... mapachichi ya ...* ("... an avocado tree and ... avocados chopped.", concept: *Parachichi* (AVOCADO), label: FALSE)

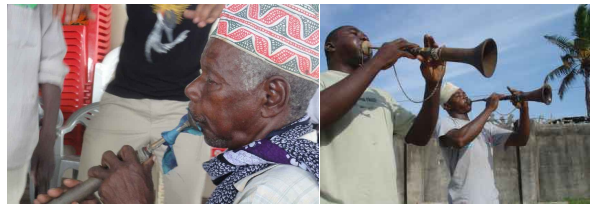

(b) *Picha ya upande wa kushoto mtu mmoja tu anapiga zumari na picha ya upande wa kulia watu wawili wanapiga zumari.* ("Picture on the left is just one person blowing the flute and in the picture on the right two people are blowing the flute.", concept: *Zumari* (FLUTE), label: TRUE)

Figure 12: More examples from MaRVL-sw.

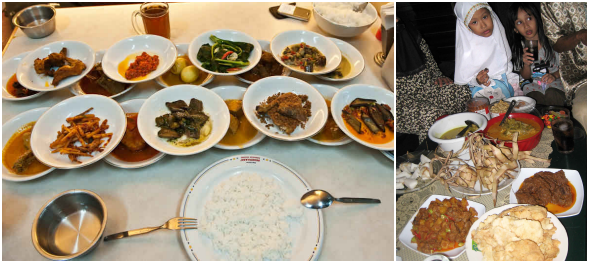

(a) *Salah satu gambar adalah gambar rendang yang disediakan di restoran Padang, dan gambar di sebelahnya adalah gambar rendang yang disajikan dengan lauk-pauk lain.* ("One of the pictures contains rendang provided at Padang restaurant, and the picture next to it contains rendang served with other side dishes.", concept: *rendang* (RENDANG, a popular Indonesian dish), label: TRUE)

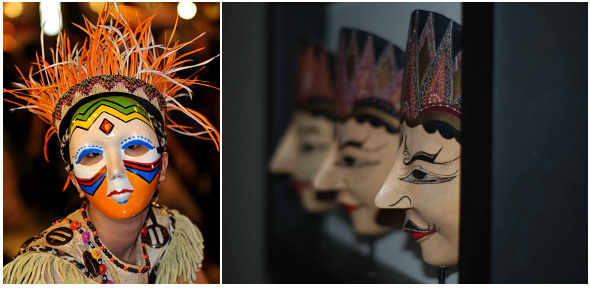

(b) *Salah satu gambar adalah topeng yang sedang dipakai seseorang, dan gambar di sebelahnya adalah gambar topeng yang dipajang.* ("In one of the pictures, someone wears a mask, and the picture next to it is a mask display.", concept: *Topeng* (MASK), label: TRUE)

Figure 13: More examples from MaRVL-ID.

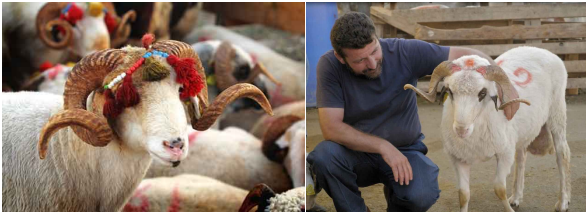

(a) *Sağdaki fotoğrafta kurban bayramı nedeniyle boynuzları süslenmiş en az iki kurbanlık hayvan bulunuyor.* ("In the right, there are at least two sacrificial animals decorated by the horns due for the sacrifice feat.", concept: *Kurban Bayramı* (EID AL-ADHA, an Islamic holiday), label: FALSE)

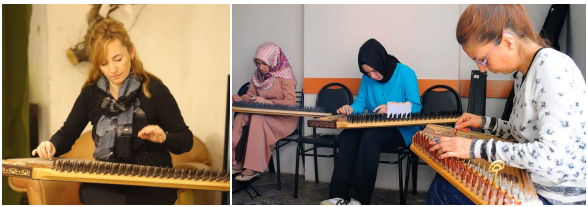

(b) *Görsellerden birinde dizlerinde kanun bulunan birden çok insan var.* ("In one of the images, there are multiple people with qanuns on their knees.", concept: *Kanun (çalgı)* (QANUN, a popular instrument in Turkey), label: TRUE)

Figure 14: More examples from MaRVL-TR.

19